\pdfoutput=1
\documentclass{article}

\usepackage{arxiv}

\providecommand{\headeright}{}
\providecommand{\undertitle}{}

\usepackage[utf8]{inputenc} % allow utf-8 input
\DeclareUnicodeCharacter{00D7}{\ensuremath{\times}}
\DeclareUnicodeCharacter{2011}{-}
\DeclareUnicodeCharacter{2013}{--}
\DeclareUnicodeCharacter{2014}{---}
\DeclareUnicodeCharacter{2019}{'}
\DeclareUnicodeCharacter{201C}{``}
\DeclareUnicodeCharacter{201D}{''}
\usepackage[T1]{fontenc}    % use 8-bit T1 fonts
\usepackage{hyperref}       % hyperlinks
\usepackage{url}            % simple URL typesetting
\usepackage{booktabs}       % professional-quality tables
\usepackage{amsfonts}       % blackboard math symbols
\usepackage{nicefrac}       % compact symbols for 1/2, etc.
\usepackage{microtype}      % microtypography
\usepackage{graphicx}
\usepackage{natbib}
\usepackage{doi}

\usepackage{algorithm}
\usepackage{algorithmic}
\usepackage{newfloat}
\usepackage{listings}
\usepackage{comment}
\usepackage{float}
\usepackage{amssymb}
\usepackage{amsmath}
\usepackage{multirow}
\usepackage[table]{xcolor}
\usepackage{makecell}
\usepackage{array}
\usepackage{tabularx}
\usepackage{caption}
\usepackage{subcaption}
\usepackage{xr}

\DeclareCaptionStyle{ruled}{labelfont=normalfont,labelsep=colon,strut=off}
\floatstyle{ruled}
\newfloat{listing}{tb}{lst}{}
\floatname{listing}{Listing}

\title{Visual Prompting Meets Feature Reconstruction-Based Anomaly Detection with Dual-Teacher Supervision}

\author{%
    \normalfont\mdseries
    Mateo Diaz-Bone, Daniel Caraballo, Florian Scheidegger, Thomas Frick, Mattia Rigotti,\\
    Andrea Bartezzaghi, Roy Assaf, Niccolo Avogaro, Yagmur G. Cinar, Brown Ebouky, Filip M. Janicki,\\
    Piotr S. Kluska, Cezary Skura, Cristiano Malossi\\
    IBM Research Europe\\
    Zurich, Switzerland\\
    \texttt{daniel.caraballo@ibm.com, eid@zurich.ibm.com}
}

\providecommand{\headeright}{A preprint}
\providecommand{\undertitle}{A preprint}
\providecommand{\shorttitle}{}

\renewcommand{\shorttitle}{Visual Prompting Meets Feature Reconstruction-Based AD}

\hypersetup{
pdftitle={Visual Prompting Meets Feature Reconstruction-Based Anomaly Detection with Dual-Teacher Supervision},
pdfsubject={Computer vision, anomaly detection},
pdfauthor={Mateo Diaz-Bone, Daniel Caraballo, Florian Scheidegger, Thomas Frick, Mattia Rigotti, Andrea Bartezzaghi, Roy Assaf, Niccolo Avogaro, Yagmur G. Cinar, Brown Ebouky, Filip M. Janicki, Piotr S. Kluska, Cezary Skura, Cristiano Malossi},
pdfkeywords={Anomaly detection, visual prompting, feature reconstruction, dual-teacher supervision, synthetic data},
}

\begin{document}
\maketitle

\begin{abstract}
Recent Anomaly Detection methods achieve perfect detection and segmentation scores on well-established datasets, such as MVTec. However, many of these methods face challenges when foundational assumptions --- such as consistent object scale, viewpoint, background, illumination, and centered placement --- are violated. Those variations that occur render anomaly detection methods unusable in many real-world scenarios. To address these limitations, we introduce three key contributions: (1) a visual prompting pipeline that isolates objects using foreground--background masking; (2) a mechanism for unfreezing the teacher in student--teacher models to improve domain adaptability; and (3) a data augmentation strategy leveraging diffusion-generated synthetic images to enhance anomaly detection performance. We achieve a 3.5 percentage point improvement over the previous state-of-the-art on the challenging AeBAD dataset by using the Masked Multiscale Reconstruction (MMR) model as our backbone.
\end{abstract}

\section{Introduction}
\label{sec:intro}
Unsupervised anomaly detection involves identifying and localizing anomalies without prior knowledge of abnormal patterns. Since the release of the first benchmark datasets in the late 2010s~\cite{nanotwice_carrera,mvtec01}, anomaly detection has received growing attention due to its broad range of applications in industrial contexts (IAD). However, as these datasets are often assembled in controlled environments, they frequently fail to capture real-world complexities, thereby limiting the practical deployment of IAD methods.

A closer look at the MVTec dataset~\cite{mvtec01} reveals that the images and their contained objects exhibit notable consistency: objects are typically aligned, centered, similarly scaled, and placed against homogeneous backgrounds. The images are also systematically captured from a fixed viewpoint under uniform illumination. While these characteristics may be enforceable in controlled industrial scenarios, numerous applications lack the ability to maintain such a sterile setup.

Recent anomaly detection methods report near-perfect detection and segmentation performance on MVTec with up to 99.9\% image level AUROC \cite{chen2024unifiedanomalysynthesisstrategy} and 99.3\% pixel level AUROC \cite{li2024industrialanomalydetectionlocalization}. However, much of this success relies on the dataset's consistency. Once these underlying assumptions are violated, performance drops significantly, as demonstrated in prior work~\cite{mmr03} and confirmed by our own experiments. This reliance on dataset-specific properties limits the real-world applicability of many IAD methods.

\begin{sloppypar}
To address these challenges, we propose three improvements for student-teacher reconstruction-based methods:
\end{sloppypar}

\begin{itemize}
    \item \textbf{Visual Prompting Masks Pipeline}: To reduce sensitivity to background variation, we design a pipeline that leverages auto-generated foreground/background masks to isolate objects and eliminate unwanted background anomalies. Applied as a postprocessing step, this enhances model robustness in the resulting scoremaps.
    
    \item \textbf{Dual Teacher Approach}: We propose a method to unfreeze the pre-trained frozen teacher in student-teacher reconstruction-based methods without suffering from model/\allowbreak{} feature collapse. This increases the model's adaptability to target domains and mitigates pre-training bias.
    
    \item \textbf{Synthetic Training Data}: We augment the training dataset with synthetic images obtained from a diffusion model. Contrary to recent methods, we generate \textit{good} images. This increases the training data diversity while at the same time acting as a regularization mechanism to prevent the model from overfitting, thereby increasing its generalization capabilities.    
\end{itemize}

The remainder is organized as follows. Section~\ref{sec:related_works} reviews related work. Section~\ref{sec:distorted_mvtec} discusses a distorted MVTec version to assess background sensitivity. Section~\ref{sec:method} details our proposed improvements. We present experiments, ablations, and results in Section~\ref{sec:experiments} and conclude in Section~\ref{sec:Conclusion}.

\begin{figure}[t]
\centering
\includegraphics[width=0.9\columnwidth]{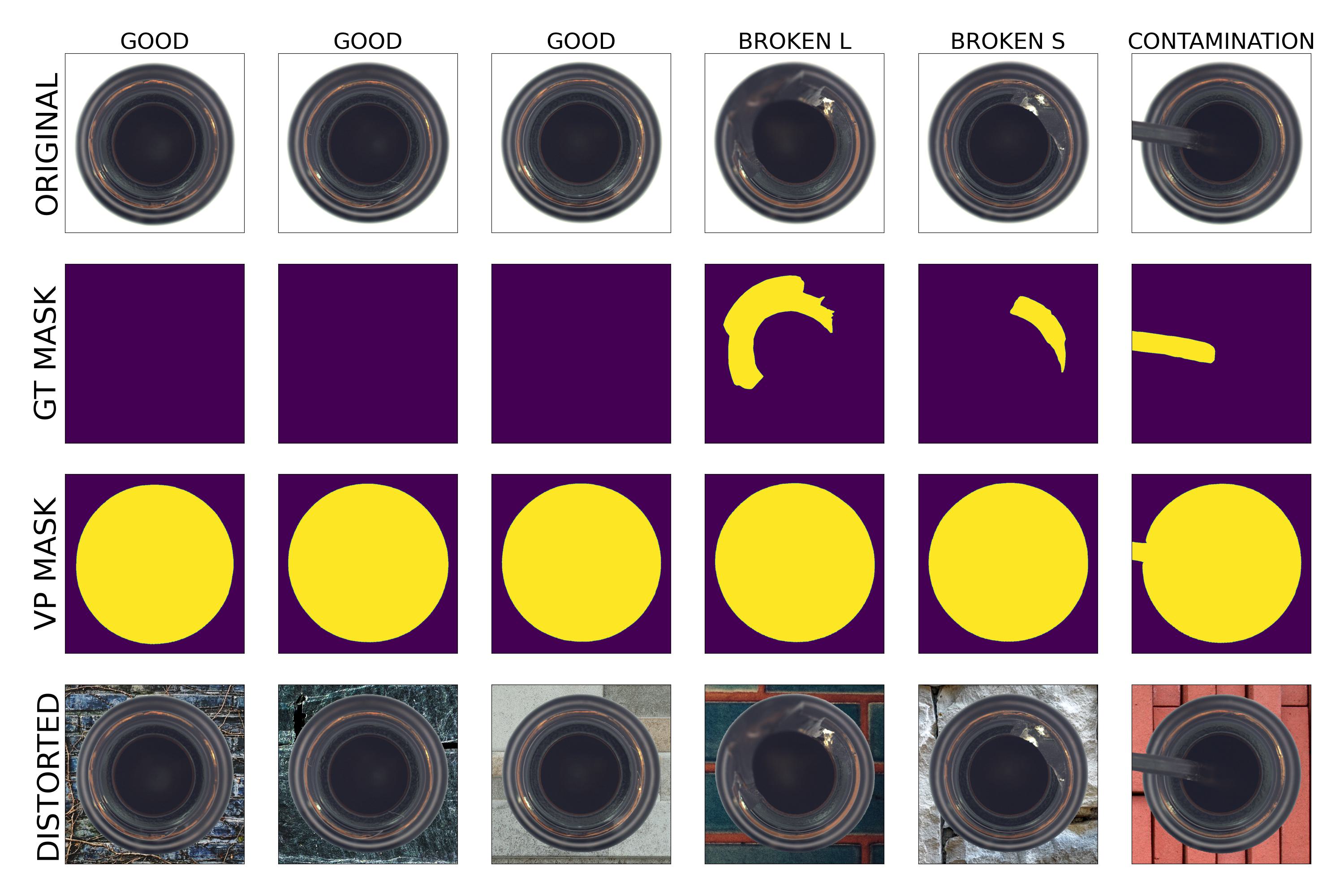}
\caption{Original and distorted MVTec~\cite{mvtec01} samples for the bottle category. We used the VP pipeline to produce human-in-the-loop annotated masks. In our distorted version the background is replaced with a textured background. 
}
\label{fig:distorted_input}
\end{figure}

\section{Related Work}
\label{sec:related_works}
Anomaly detection (AD) for visual data has gained significant attention, particularly in unsupervised anomaly detection (UAD) settings where models must perform without access to defective samples during training. Traditional datasets like MVTec AD~\cite{mvtec01} and VisA~\cite{visa02} have been extensively used to benchmark models in this field. However, these datasets often assume strong spatial alignment and lack the variability needed to assess model robustness under domain shifts. To address this limitation, the AeBAD~\cite{mmr03} dataset was introduced, offering diverse domain shifts across defect categories and providing a more comprehensive evaluation of AD models under real-world conditions with varying distributions. The MMR model~\cite{mmr03} was proposed to handle such domain shifts. MMR utilizes multi-scale features for reconstructing normal patterns, effectively detecting anomalies by highlighting reconstruction errors. Despite its strengths, MMR still relies on a frozen pre-trained encoder, limiting its adaptability to new domains.

Further, AD methodologies have predominantly relied on one-class learning, where models are trained exclusively on normal samples to generalize to unseen defective cases during inference. Memory-based models, such as PatchCore~\cite{PatchCore04}, GraphCore~\cite{GraphCore05}, and FAPM~\cite{FAPM06}, store nominal features for later comparison, achieving impressive results. While effective, these methods may not adapt well under significant domain shifts due to their reliance on stored normal features.

To enhance feature space separation and improve discriminative power, contrastive learning and self-supervised techniques have been employed in AD tasks~\cite{Draem09, Cutpaste10,DCdetector17}
where methods like ReConPatch~\cite{Reconpatch08} encourage models to learn more robust and discriminative representations, facilitating better AD. However, they often require careful design of pretext tasks and may still struggle when anomalies exhibit significant differences from normal data.

Knowledge distillation techniques, where a frozen teacher model guides a student in learning representations, are widely used~\cite{DeSTSeg11}. Methods like Reverse Distillation~\cite{ReverseDistillation12} and EfficientAD~\cite{EfficientAD07} ensure robust feature extraction but limit domain adaptability due to the frozen teacher. Our approach addresses this by unfreezing the teacher within a dual-teacher framework, enhancing adaptability while preventing feature collapse. Concurrent works such as Recontrast~\cite{guo2023recontrastdomainspecificanomalydetection} attempt a related type of supervision which is specific to the Reverse Distillation framework. This makes it less flexible to incorporate into other distillation frameworks.

Recently, diffusion models~\cite{DDPM13, LatentDiffusion14} have been used to generate synthetic anomalies, turning unsupervised AD into a supervised task~\cite{RealNet16, AnoDDPM15}. However, creating relevant synthetic defects and integrating them without bias remains challenging. Instead, we synthesize high-quality normal images to enhance dataset diversity in the target domain.

While isolating objects from the background can improve detection, it has remained largely underexplored in UAD. As far as we know, the only related work~\cite{baugh2023zeroshotanomalydetectionpretrained}, tiles multi-object images into single objects through foreground extraction. We extend this field by using Visual Prompting (Section~\ref{subsec:VP_BG_Removal}) to create segmentation masks that enhance anomaly detection.

\section{A Distorted Version of MVTec}
\label{sec:distorted_mvtec}

%%%%%%%%%%%%%%%%%%%%%%%%%%%%%%%%%%%%%%%%%% FIGURE

%%%%%%%%%%%%%%%%%%%%%%%%%%%%%%%%%%%%%%%%%% FIGURE

The CC~BY-NC-SA~4.0\footnote{\url{https://creativecommons.org/licenses/by-nc-sa/4.0/}} licensed MVTec~\cite{mvtec01} remains a key AD benchmark but relies on controlled, aligned samples, limiting real-world applicability. Inspired by Fashion MNIST~\cite{xiao2017fashion} as a drop-in replacement for MNIST~\cite{deng2012mnist}, we introduce a distorted MVTec variant that retains the same splits, defect categories, and masks, but replaces backgrounds with randomly sampled textures from high-resolution, CC-licensed\footnote{\url{https://creativecommons.org/publicdomain/}} images stemming from the public domain. Details about the data are in Appendix~E.
% \ref{app:data_contributions} == E

Figure~\ref{fig:distorted_input} shows examples from the bottle category, including human-validated segmentation masks that help create more realistic blended samples.  

We evaluate popular AD methods using anomalib\footnote{\url{https://github.com/openvinotoolkit/anomalib}} ~\cite{akcay2022anomalib} on both original and distorted MVTec. Performance drops by 1.7\%-points up to 16\%-points on the distorted dataset, with background noise especially noticeable in image corners, sometimes altering the ranking of algorithms. Table~\ref{tab:mergedistorted} quantifies these effects and confirms the necessity of reevaluating AD methods in less controlled conditions. Additional visual examples comparing the performance of different algorithms are provided in Appendix~A. Our proposed two-stage pipeline, where visual prompting (VP) identifies the problem-specific region of interest, followed by a more traditional AD pipeline, mitigates this limitation.

\begin{table*}[ht]
\centering
\begin{small} % EID to make it fit.
\begin{tabular}{lrrrrrrrrrrrr}
\toprule
Algorithm & \multicolumn{2}{c}{Fre} & \multicolumn{2}{c}{Padim} & \multicolumn{2}{c}{PatchCore} & \multicolumn{2}{c}{RD} & \multicolumn{2}{c}{Uflow} & \multicolumn{2}{c}{MMR} \\
Data & Ref & Dist & Ref & Dist & Ref & Dist & Ref & Dist & Ref & Dist & Ref & Dist \\
\midrule
bottle     & 97.4 & 88.1 & 97.8 & 97.8 & 98.2 & 96.1 & 98.7 & 90.5 & 95.7 & 97.4 & 98.67 & 93.17 \\
cable      & 97.2 & 89.8 & 95.1 & 92.8 & 98.1 & 95.3 & 96.5 & 87.2 & 98.4 & 97.1 & 95.82 & 84.85 \\
capsule    & 97.8 & 70.7 & 98.2 & 87.3 & 98.8 & 81.2 & 98.8 & 69.0 & 98.8 & 98.0 & 97.75 & 65.01 \\
hazelnut   & 98.3 & 93.3 & 96.9 & 93.5 & 98.5 & 97.2 & 98.7 & 91.0 & 99.1 & 98.9 & 98.98 & 94.76 \\
metal\_nut & 97.3 & 80.8 & 95.0 & 86.8 & 98.2 & 91.3 & 96.6 & 77.7 & 97.7 & 96.4 & 95.24 & 62.78 \\
pill       & 96.3 & 68.4 & 95.5 & 86.8 & 97.6 & 85.2 & 97.2 & 63.1 & 99.2 & 98.6 & 98.55 & 73.42 \\
screw      & 96.3 & 66.4 & 97.7 & 77.1 & 98.9 & 84.3 & 99.4 & 82.6 & 99.3 & 93.5 & 99.49 & 82.85 \\
toothbrush & 98.1 & 84.2 & 98.7 & 76.5 & 98.6 & 89.8 & 99.0 & 91.4 & 98.0 & 97.7 & 98.63 & 88.89 \\
transistor & 98.5 & 89.5 & 97.0 & 89.4 & 97.1 & 77.9 & 91.3 & 73.6 & 97.7 & 86.6 & 91.27 & 73.61 \\
zipper     & 95.5 & 77.0 & 96.8 & 95.1 & 97.9 & 92.5 & 97.9 & 82.5 & 97.1 & 98.4 & 97.95 & 85.33 \\
\midrule
mean       & 97.3 & 80.8 & 96.9 & 88.3 & 98.2 & 89.1 & 97.4 & 80.9 & 98.1 & 96.3 & 97.24 & 80.47 \\
Change     & \multicolumn{2}{c}{-16.4} & \multicolumn{2}{c}{-8.6} & \multicolumn{2}{c}{-9.1} & \multicolumn{2}{c}{-16.6} & \multicolumn{2}{c}{-1.8} & \multicolumn{2}{c}{-16.7} \\
\bottomrule
\end{tabular}
\end{small}
\caption{Performance gap between the reference and distorted versions on the MVTec dataset. All reference results are reproduced with default setups for a fair comparison. The background distortion leads to a challenging scenario that negatively impacts traditional AD algorithms. References: Fre~\cite{ndiour2022fre}, Padim~\cite{defard2021padim}, PatchCore~\cite{PatchCore04}, Reverse Distillation (RD)~\cite{ReverseDistillation12}, Uflow~\cite{tailanian2024uflow}, and MMR~\cite{mmr03}.}
\label{tab:mergedistorted}
\end{table*}

\section{Method}
\label{sec:method}
\begin{figure}[t]
\centering
% ---------- Top figure ----------
\begin{minipage}{\columnwidth}
    \centering
    \includegraphics[width=0.9\linewidth]{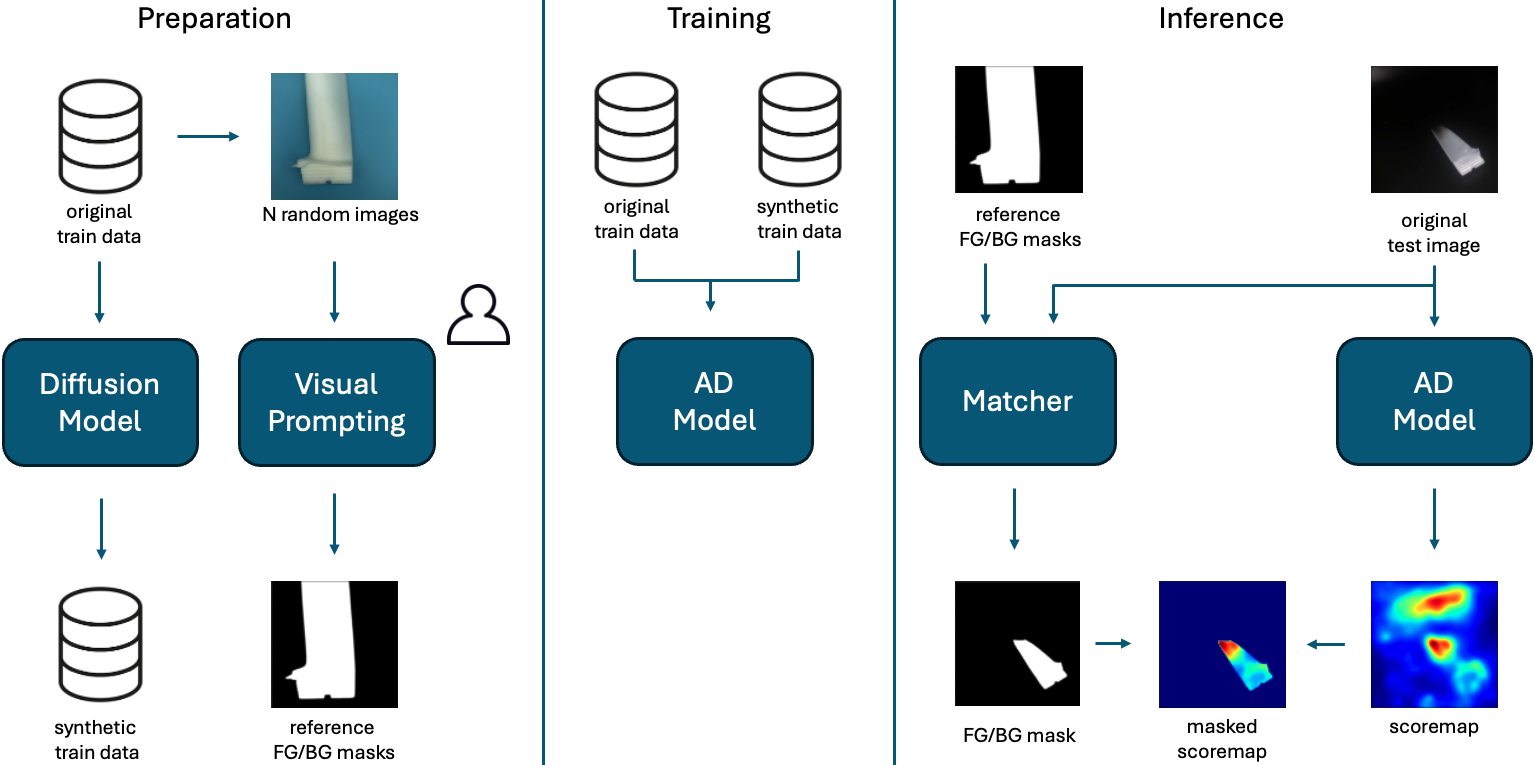}
    \caption{Overview of our proposed framework.}
    \label{fig:diagram_pipeline}
\end{minipage}

\vspace{1.5em}  % vertical space between figures
% ---------- Bottom figure ----------
\begin{minipage}{\columnwidth}
    \centering
    \includegraphics[width=0.9\linewidth]{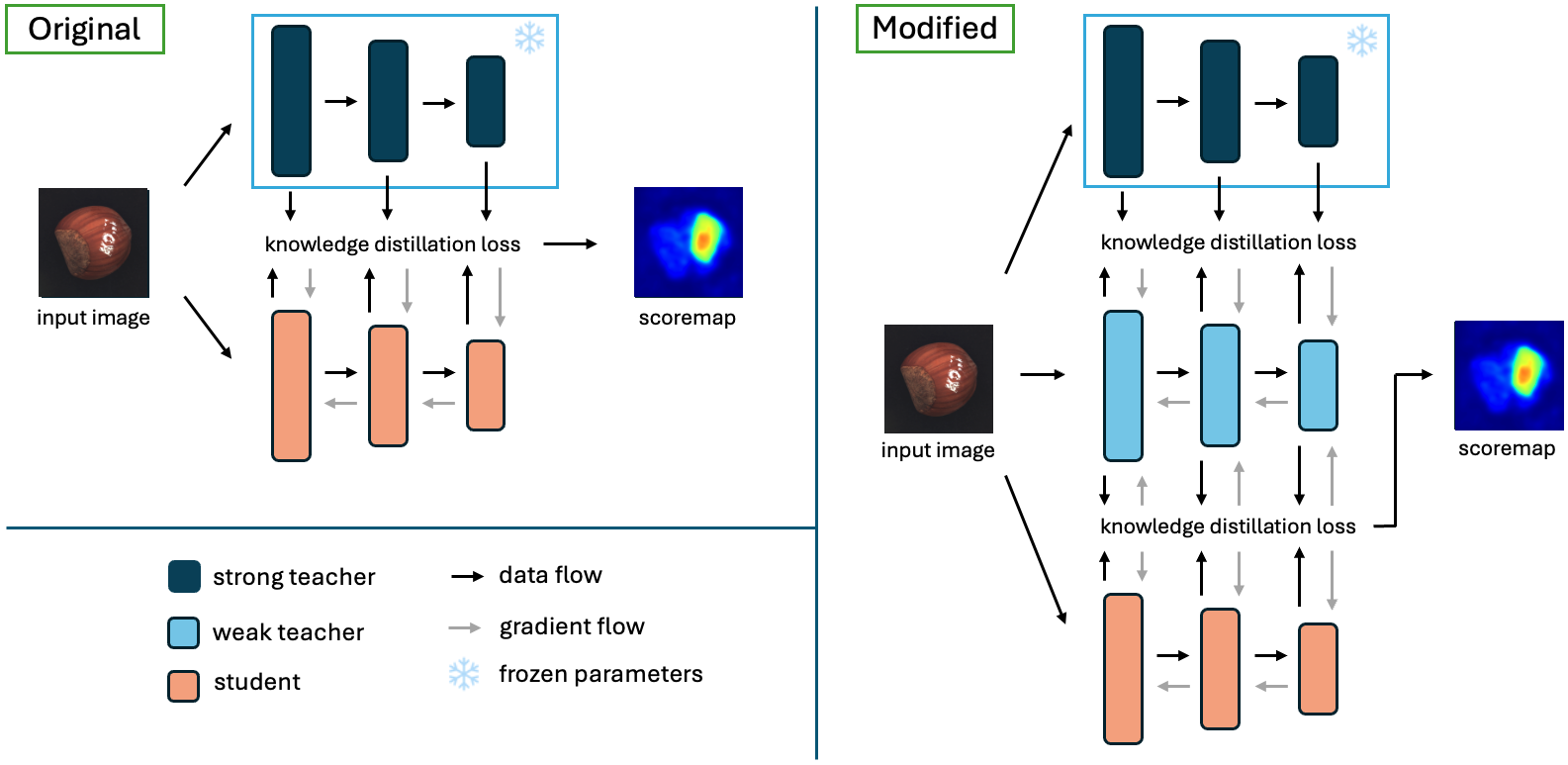}
    \caption{Original and modified version of student-teacher feature reconstruction (MMR-style).}
    \label{fig:diagram_3way_MMR}
\end{minipage}
\end{figure}
%%%%%%%%%%%%%%%%%%%%%%%%%%%%%%%%%%%%%%%%%% FIGURE

In this section, we present our novel framework for anomaly detection, which features synthetic data generation, a knowledge-preserving dual-teacher approach, and visual prompting. Our framework is architecture-agnostic and can be applied to most knowledge-distillation-based reconstruction methods. An overview of the proposed pipeline is illustrated in Figure~\ref{fig:diagram_pipeline}. The pipeline is divided into three phases:

\textbf{Preparation.}
Before training begins, the original training data is used to train a diffusion model, which is then used to sample synthetic images that resemble the original data (Section~\ref{subsec:diffusion_model}). In parallel, the user selects a small number of random images from the training set and creates foreground/background (FG/BG) masks using Visual Prompting (Section~\ref{subsec:VP_BG_Removal}).

\textbf{Training.}
During training, the model learns from a combination of real and synthetic training data.

\textbf{Inference.}
During inference, the test image is passed through the model to generate an intermediate score map. In parallel, the test image is also processed by the Matcher (Section~\ref{subsec:VP_BG_Removal}), which uses the reference FG/BG masks from the preparation phase to automatically generate a FG/BG mask for the test image. This mask is then applied to the score map to produce the final output.

\subsection{Synthetic Train Data via Diffusion Models}\label{subsec:diffusion_model}
%Inspired by RealNet \cite{RealNet16}, we employ a pre-trained diffusion model to generate synthetic training data. By fine-tuning the diffusion model with the training dataset, we can sample synthetic images which resemble original 'good' train images from the target domain. We use those sampled 'good' images to augment the training dataset. The synthetic images are combined with real images to enrich the training set and provide diverse examples.

Inspired by RealNet~\cite{RealNet16}, we employ a pre-trained diffusion model to generate synthetic training data. By fine-tuning the diffusion model on the target domain's training set, we sample synthetic images that closely resemble the original good training images in both appearance and context. These synthetic samples are then added to the training set alongside the real images, enriching it with diverse, domain-relevant examples.

The inherent stochasticity of the diffusion process further enhances data diversity, enabling the model to better capture the domain's data distribution. Additionally, the synthetic images act as a form of regularization, mitigating overfitting to the original training data and improving generalization.

\subsection{Knowledge-Preserving Dual Teacher Approach}
Knowledge Distillation (KD) based student-teacher feature reconstruction architectures have received much attention in recent years due to their outstanding performances.
% A common configuration is depicted in Figure \ref{fig:diagram_3way_MMR} (left).
Figure~\ref{fig:diagram_3way_MMR} (left) depicts a common configuration as featured in MMR~\cite{mmr03}.
In this method, the student tries to reconstruct the features obtained from a frozen teacher network which was pre-trained on a large-scale dataset like Imagenet~\cite{imagenet}. 
However, AD images often differ significantly from the natural images used in large-scale pre-training datasets, making the pre-trained encoder features less suitable for the target domain and restricting performance. Since the teacher encoder is typically frozen, it cannot adapt these features to the target domain.

To circumvent this structural limitation we present a method to unfreeze the teacher. If one naively unfreezes the teacher encoder, the model would inevitably suffer from feature collapse as it rapidly converges to a trivial solution to minimize the student teacher distance. To prevent this, we introduce a secondary teacher which acts as a regularization for the unfrozen teacher. Figure~\ref{fig:diagram_3way_MMR} (right) depicts our proposed modified architecture.

We denote the unfrozen teacher as 'weak teacher' and the supervisory teacher as 'strong teacher'. Both teachers are instantiated with the same pre-trained weights. To control the level of flexibility of the weak teacher we modify the loss function.
\begin{equation}
\mathcal{L}_{\text{total}} = \mathcal{L}_{\text{S\_WT}} + \lambda \mathcal{L}_{\text{WT\_ST}}
\end{equation}

Here 
$\mathcal{L}_{\text{S\_WT}}$ 
denotes the KD loss between the student and the weak teacher and $\mathcal{L}_{\text{WT\_ST}}$ 
denotes the KD loss between the weak teacher and the strong teacher. 
The type of KD loss stays consistent with the original KD loss function which is cosine similarity inherited from the original formulation of MMR. The hyperparameter $\lambda$ balances the two losses, controlling the regularization strength of the strong teacher. High $\lambda$ values limit the weak teacher's flexibility to deviate from its original parameters, while low values allow more adaptation to the student. The goal is to set $\lambda$ low enough to enable adaptation without causing feature collapse. During inference, the strong teacher is disabled to reduce inference time, as it does not contribute to scoremap calculation.

\begin{table*}[!ht]
\centering

% ---------- (a) Pixel-level ----------
\begin{minipage}{\textwidth}
\centering

\begin{tabular}{lccccc}
\toprule
Method (Cited Results) & Same & Background & Illum.\ & View & Mean \\
\midrule
PatchCore           & 89.5 ± 0.2 & 89.4 ± 0.1 & 88.2 ± 0.1 & 84.0 ± 0.2 & 87.8 \\
ReverseDistillation & 86.4 ± 0.4 & 86.4 ± 0.7 & 86.7 ± 0.5 & 82.9 ± 0.7 & 85.6 \\
DRAEM               & 71.4 ± 4.2 & 44.3 ± 11.6 & 67.6 ± 2.7 & 71.1 ± 2.3 & 63.6 \\
NSA                 & 43.0 ± 1.3 & 29.7 ± 2.1 & 59.9 ± 1.3 & 51.1 ± 0.1 & 45.9 \\
RIAD                & 71.9 ± 1.3 & 33.4 ± 0.6 & 65.3 ± 1.0 & 62.2 ± 1.7 & 58.2 \\
InTra               & 76.8 ± 0.2 & 74.8 ± 0.3 & 73.7 ± 0.3 & 73.4 ± 0.2 & 74.7 \\
MMR                 & 89.6 ± 0.2 & 90.1 ± 0.2 & 90.2 ± 0.2 & 86.3 ± 0.3 & 89.1 \\
MiniMaxAD           & 78.5 ± 0.2 & 79.4 ± 0.4 & 83.7 ± 0.3 & 76.3 ± 0.3 & 79.4 \\
\midrule
\textbf{MMR+++ (Ours)} & \textbf{91.3} ± 0.0 & \textbf{91.1} ± 0.0 & \textbf{91.3} ± 0.1 & \textbf{86.9} ± 0.1 & \textbf{90.2} \\
\bottomrule
\end{tabular}

\captionof{table}{Pixel-level anomaly detection (AUPRO) on \textsc{AeBAD-S}. Bold denotes the best results.}
\label{tab:aebadaupro}
\end{minipage}

\vspace{1.5em} % vertical gap between the two tables

% ---------- (b) Image-level ----------
\begin{minipage}{\textwidth}
\centering
\begin{tabular}{lccccc}
\toprule
Method (Cited Results) & Same & Background & Illum.\ & View & Mean \\
\midrule
PatchCore           & 75.2 ± 0.3 & 74.1 ± 0.3 & 74.6 ± 0.4 & 60.1 ± 0.4 & 71.0 \\
ReverseDistillation & 82.4 ± 0.6 & 84.3 ± 0.9 & 85.5 ± 0.9 & 71.9 ± 0.8 & 81.0 \\
DRAEM               & 64.0 ± 0.4 & 62.1 ± 6.1 & 61.6 ± 2.7 & 62.3 ± 0.9 & 62.5 \\
NSA                 & 66.5 ± 1.4 & 48.8 ± 3.5 & 55.5 ± 3.2 & 55.9 ± 1.1 & 56.7 \\
RIAD                & 38.6 ± 0.6 & 41.6 ± 1.3 & 46.8 ± 0.8 & 33.0 ± 0.6 & 40.0 \\
InTra               & 39.8 ± 0.8 & 46.1 ± 0.5 & 44.7 ± 0.3 & 46.3 ± 1.5 & 44.2 \\
MMR                 & 85.6 ± 0.5 & 84.4 ± 0.7 & 88.8 ± 0.5 & 79.9 ± 0.6 & 84.7 \\
MiniMaxAD           & 86.3 ± 0.5 & 89.0 ± 0.4 & 85.9 ± 0.7 & \textbf{86.9} ± 0.5 & 87.0 \\
\midrule
\textbf{MMR+++ (Ours)} & \textbf{88.5} ± 0.3 & \textbf{90.6} ± 0.2 & \textbf{91.5} ± 0.3 & 82.1 ± 0.4 & \textbf{88.2} \\
\bottomrule
\end{tabular}
\captionof{table}{Image-level anomaly detection (I-AUROC) on \textsc{AeBAD-S}. Bold denotes the best results.}
\label{tab:aebadauroc}

\end{minipage}

\end{table*}

\subsection{Background Removal via Visual Prompting}\label{subsec:VP_BG_Removal}

To enhance prediction robustness in images with irregular backgrounds, we introduce a pipeline that generates FG/BG segmentation masks, isolating the object of interest by masking out the background. A diagram of the pipeline is shown in Figure~\ref{fig:diagram_pipeline}. The process begins by selecting a few reference images and using Visual Prompting with the Segment Anything Model (SAM)~\cite{kirillov2023segment} to create initial FG/BG segmentation masks. During inference, the Matcher framework~\cite{liu2024matchersegmentshotusing}, built on the scalable feature-matching approach by Frick et al.~\cite{frick2024interactive}, auto-generates masks for query images by referencing these user-generated masks, enabling efficient, high-quality mask production across datasets with minimal human input. This method further benefits from DinoV2~\cite{oquab2024dinov2learningrobustvisual} to extract meaningful features for feature matching, enhancing the precision of the Matcher pipeline. The masks are then applied to the model's score maps as a postprocessing step, without requiring model retraining.

\section{Experiments}
\label{sec:experiments}

In this section, we evaluate our proposed framework. For our experiments, we use the MMR~\cite{mmr03} model as our backbone. We perform extensive experiments on the CC~BY~4.0-licensed\footnote{\url{https://creativecommons.org/licenses/by/4.0/}} AeBAD-S~\cite{mvtec01} dataset which entails many of the challenges that we are tackling with our improvements. 

% For completeness, we validate our improvements on the MVTec dataset to demonstrate that our framework improves the predictions even in simpler settings.
% improves the predictions -> operates reasonably 
We validate our approaches on the MVTec dataset to demonstrate that our framework operates reasonably even in highly controlled environments that are missing the fundamental challenges that we are addressing. % simpler settings.

\subsection{Experimental Settings}
\label{subsect:experiment_implmentation}

\textbf{Datasets.}
We evaluate on two datasets. 
\textbf{AeBAD-S} includes 521 training images and 1639 testing images of aero-engine blades, captured under varying lighting conditions and backgrounds. 
\textbf{MVTec} encompasses 15 sub-datasets (10 objects, 5 textures) totaling 5354 images.

\textbf{Synthetic Augmentation.}
We adopt the DDPM approach from \cite{RealNet16} to sample additional training images by setting the perturbation strength parameter to 0. For \textbf{AeBAD-S}, we add 3$\times$ more training images; for \textbf{MVTec}, we maintain a 1:1 ratio between real and synthetic images.

\textbf{Model Training.}
We primarily follow the training recipe of MMR~\cite{mmr03}, scaling batch size according to the amount of synthetic data. For real images, we use the original resizing/cropping augmentations; for synthetic images, we apply center-cropping and horizontal flips. We set $\lambda=1.5$ for \textbf{AeBAD-S} and $\lambda=0.5$ for \textbf{MVTec}.

\textbf{Visual Prompting.}
For SAM~\cite{kirillov2023segment}, we use a single (AeBAD-S) or five (MVTec) reference FG/BG masks. We employ a DINOv2 ViT-L backbone in Matcher and dilate the FG/BG masks (40 pixels for AeBAD-S, 15 pixels for MVTec) to account for object borders.

%\input{tabs/merged_aebad}

% \subsection{Results}
\section{Results}
\label{sec:results}

To evaluate the image level predictions we use the area under the receiver operating characteristic (I-AUROC) which captures the relation between true positives and false positives. For anomaly segmentation, we use pixel level Auroc P-AUROC and the area under the per-region-overlap (AUPRO) as the evaluation metric which normalizes anomalies by their size to treat anomalies of any size equally.

\textbf{MVTec} The results in Table~\ref{tab:config_ablation_v2} underscore the effectiveness of background removal and the dual teacher on the MVTec dataset. When comparing to the baseline configuration (Config A), we note a progressive improvement in all three performance metrics as we iteratively add components. It is important to note, that masking was excluded from the Transistor case, due to a conceptional limitation that exists between background masking and logical defects. 

Synthetic data, however, was found to impair performance on the MVTec dataset. Inspection reveals that many synthetic images labeled as ``good'' appeared ``defect-like,'' likely causing these degraded results. Additionally, since training data directly influences the unfrozen teacher model, misaligned synthetic data negatively impacts teacher fine-tuning. This explains the consistent declines across all three metrics when synthetic data is combined with the dual teacher.

\textbf{AeBAD-S}
The results are presented in Table~\ref{tab:aebadaupro} and Table~\ref{tab:aebadauroc}, which are averaged over 5 runs (3 for MiniMaxAD). We compare our method against established baselines on both pixel-level and image-level anomaly detection tasks on \textsc{AeBAD-S}. Baselines include PatchCore~\cite{PatchCore04} (CVPR'22), ReverseDistillation~\cite{ReverseDistillation12} (CVPR'22), DRAEM~\cite{Draem09} (ICCV'21), NSA~\cite{NSAschluter2022naturalsyntheticanomaliesselfsupervised} (ECCV'22), RIAD~\cite{riadZAVRTANIK2021107706} (PR'20), InTra~\cite{intrapirnay2021inpaintingtransformeranomalydetection} (ICIAP'22), MMR~\cite{mmr03} (CII'23), and MiniMaxAD~\cite{wang2024minimaxadlightweightautoencoderfeaturerich}.

Our configuration achieves an I-AUROC of \textbf{88.2\%}, exceeding the previous SOTA of 87.0\% by a remarkable \textbf{1.2\%}. Compared to MMR, which is the backbone used in our configuration, our proposed framework increases the I-AUROC by a staggering \textbf{3.5\%}. At the same time, our configuration produces an AUPRO score of \textbf{90.2\%} which exceeds the previously reported SOTA of 89.1\% by \textbf{1.1\%}.

\subsection{Hyperparameter Ablations}
All hyperparameter ablations are performed on AeBAD-S which covers the challenges that our improvements are designed for. The base model used is MMR \cite{mmr03}.
\label{app:hyperparameter_ablations}
\subsubsection{Synthetic Data}
%%%%%%%%%%%%%%%%%%%%%%%%%%% MERGED TABLES
% This table contains all information. REF + DIST 
% And mean average, mean average change. 

\begin{table*}[h]
\centering
% \begin{tabular}{|c|c|c|c|c|c|c|}
\begin{tabular}{lrrrrr}
\toprule
Data R|S     & b = 16 (1x)  & b = 32 (2x) & b = 48 (3x) & b = 64 (4x) & b = 80 (5x)  \\ 
\midrule
0 | 521  & 80.3 / 87.6  & 79.9 / 87.3 & --- / --- & 79.5 / 87.3 & --- / ---  \\ 
521 | 0   & \textbf{84.7 / 88.6}  & 84.6 / 88.5 & 84.0 / 88.5 & 83.5 / 88.1 & 83.2 / 88.3   \\ 
521 | 521 & 83.5 / 87.7  & \textbf{85.7 / 88.6}  & 85.1 / 88.6 & 84.6 / 88.3  & 84.0 / 88.4  \\ 
521 | 1042 & 81.7 / 87.1  & 85.5 / 88.7  & 85.6 / 88.9 & 85.3 / 88.5  & 85.1 / 88.6   \\ 
521 | 1563 & 80.1 / 85.5  & 84.9 / 88.7  & \textbf{86.0} / \textbf{88.9} & 85.7 / 88.7 & 85.5 / 88.7  \\ 
521 | 2084  & 78.5 / 83.0  & 83.5 / 87.1  & 85.4 / 88.8 & \textbf{85.7 / 88.9}  & \textbf{85.4 / 88.8} \\ 
521 | 3647  & 76.7 / 81.1  & 79.4 / 83.6 & 82.4 / 87.0 & 84.5 / 88.2 & 85.2 / 88.8   \\ 
\bottomrule
\end{tabular}
\caption{Ablation study for AeBAD-S real (R) and synthetic (S) data on MMR. Evaluation metrics are Image AUROC / AUPRO. b denotes batch size. Results are averaged over three runs.}
\label{table:synth_ablation_aebad}
\end{table*}

In Table~\ref{table:synth_ablation_aebad}, we ablate the effects of synthetic data volume and batch size. The optimal batch size scales roughly linearly with the total data. The best performance is achieved with 1563 synthetic images (a 1:3 real-to-synthetic ratio) and a batch size of 48 (3$\times$ the original). Notably, the best configurations for I-AUROC and AUPRO coincide. Even with purely synthetic training data, the model outperforms most baselines.

\subsubsection{Dual Teacher}

\begin{table*}[!h]
\centering
\begin{small}
\begin{tabular}{l|ccc|ccc|cc}
\toprule
% & \multirow{2}{*}{Syn. Data} & \multirow{2}{*}{Dual Teach} & \multirow{2}{*}{BG-Rem} & \multicolumn{3}{|c|}{MVTecAD} & \multicolumn{2}{|c}{AeBAD-S} \\
% \cmidrule{5-9}
% & & & & I-AUROC & P-AUROC & AUPRO & I-AUROC & AUPRO \\

& \multirow{2}{*}{Syn.} & \multirow{2}{*}{Dual} & \multirow{2}{*}{BG} & \multicolumn{3}{|c|}{MVTecAD} & \multicolumn{2}{|c}{AeBAD-S} \\
\cmidrule{5-9}
& Data & Teach & Rem & I-AUROC & P-AUROC & AUPRO & I-AUROC & AUPRO \\
\midrule
A &  &  &  & 98.3 ± 0.0 & 97.2 ± 0.0 & 92.7 ± 0.0 & 84.7 ± 0.1 & 88.7 ± 0.1 \\ 
B & \checkmark &  &  & {98.4} ± 0.0 & {97.1} ± 0.0 & {92.3} ± 0.0 & 86.0 ± 0.1 & 88.9 ± 0.0 \\
C & & \checkmark &  & 98.4 ± 0.0 & 97.6 ± 0.0 & {92.9} ± 0.0 & 85.7 ± 0.2 & 89.3 ± 0.1 \\ 
D & & & \checkmark & 98.5 ± 0.0 & {97.3} ± 0.0 & 93.0 ± 0.0 & 85.8 ± 0.1 & 89.3 ± 0.0 \\ 
E & &\checkmark &\checkmark & 98.6 ± 0.0 & \textbf{97.7} ± 0.0 & \textbf{93.2} ± 0.0 & 86.8 ± 0.3 & 89.9 ± 0.1 \\
F & \checkmark & \checkmark & & {98.0} ± 0.2 & {97.1} ± 0.1 & {92.0} ± 0.2 & 87.2 ± 0.2 & 89.7 ± 0.1 \\
G & \checkmark &  & \checkmark & \textbf{98.7} ± 0.0 & {97.3} ± 0.0 & {92.8} ± 0.0 & 86.6 ± 0.1 & 89.4 ± 0.0 \\
H & \checkmark & \checkmark & \checkmark & {98.6} ± 0.0 & {97.4} ± 0.0 & {92.9} ± 0.0 & \textbf{88.2} ± 0.1 & \textbf{90.2} ± 0.1 \\ 
\bottomrule
\end{tabular}
\end{small}
\caption{Ablation on MVTec AD and AeBAD-S. Results are reported from the last epoch. For AeBAD-S the results are averaged over 5 runs.}
\label{tab:config_ablation_v2}
\end{table*}

\begin{table*}[!h]
\centering
% \begin{tabular}{|c|c|c|c|c|}
\begin{tabular}{ccccc}
\toprule
Reference Images     & Pairwise mIoU  & HITL mIoU & I-AUROC & AUPRO \\
\midrule
1 & 97.7 ± 0.7 & 98.3 ± 0.6 & 88.2 ± 0.1   & 90.2 ± 0.1  \\
5 & 98.8 ± 0.1 & 98.7 ± 0.1 & 88.2 ± 0.1   & 90.2 ± 0.1  \\
15 & 99.0 ± 0.2 & 98.9 ± 0.2 & 88.2 ± 0.1   & 90.1 ± 0.1  \\
\bottomrule
\end{tabular}
\caption{Ablation study on the number of reference images given to Matcher for AeBAD-S dataset. Results are averaged over three different FG/BG mask sets using N random images as references. 'Pairwise mIoU' denotes the average mIoU in between different Matcher generated FG/BG mask sets. 'HITL mIoU' denotes the average mIoU in between the Matcher generated FG/BG mask sets and the human-in-the-loop generated FG/BG mask set.}
\label{table:mask_iou_exps}
\end{table*}

% \begin{table*}[ht]
% \centering
% \begin{tabular}{|c|c|c|c|c|}
% \hline
% Reference Images     & Pairwise mIoU  & HITL mIoU & I-AUROC & AUPRO \\ \hline
% 1 & 97.7 ± 0.7 & 98.3 ± 0.6 & 88.2 ± 0.1   & 90.2 ± 0.1  \\ \hline
% 5 & 98.8 ± 0.1 & 98.7 ± 0.1 & 88.2 ± 0.1   & 90.2 ± 0.1  \\ \hline
% 15 & 99.0 ± 0.2 & 98.9 ± 0.2 & 88.2 ± 0.1   & 90.1 ± 0.1  \\ \hline
% \end{tabular}
% \caption{Ablation study on the number of reference images given to Matcher for AeBAD-S dataset. Results are averaged over three different FG/BG mask sets using N random images as references. 'Pairwise mIoU' denotes the average mIoU in between different Matcher generated FG/BG mask sets. 'HITL mIoU' denotes the average mIoU in between the Matcher generated FG/BG mask sets and the human-in-the-loop generated FG/BG mask set.}
% \label{table:mask_iou_exps}
% \end{table*}

\begin{figure*}[!ht]
\centering

% ---------- Left (a): AUROC ----------
\begin{minipage}{0.48\textwidth}
    \centering
    \includegraphics[width=0.9\linewidth]{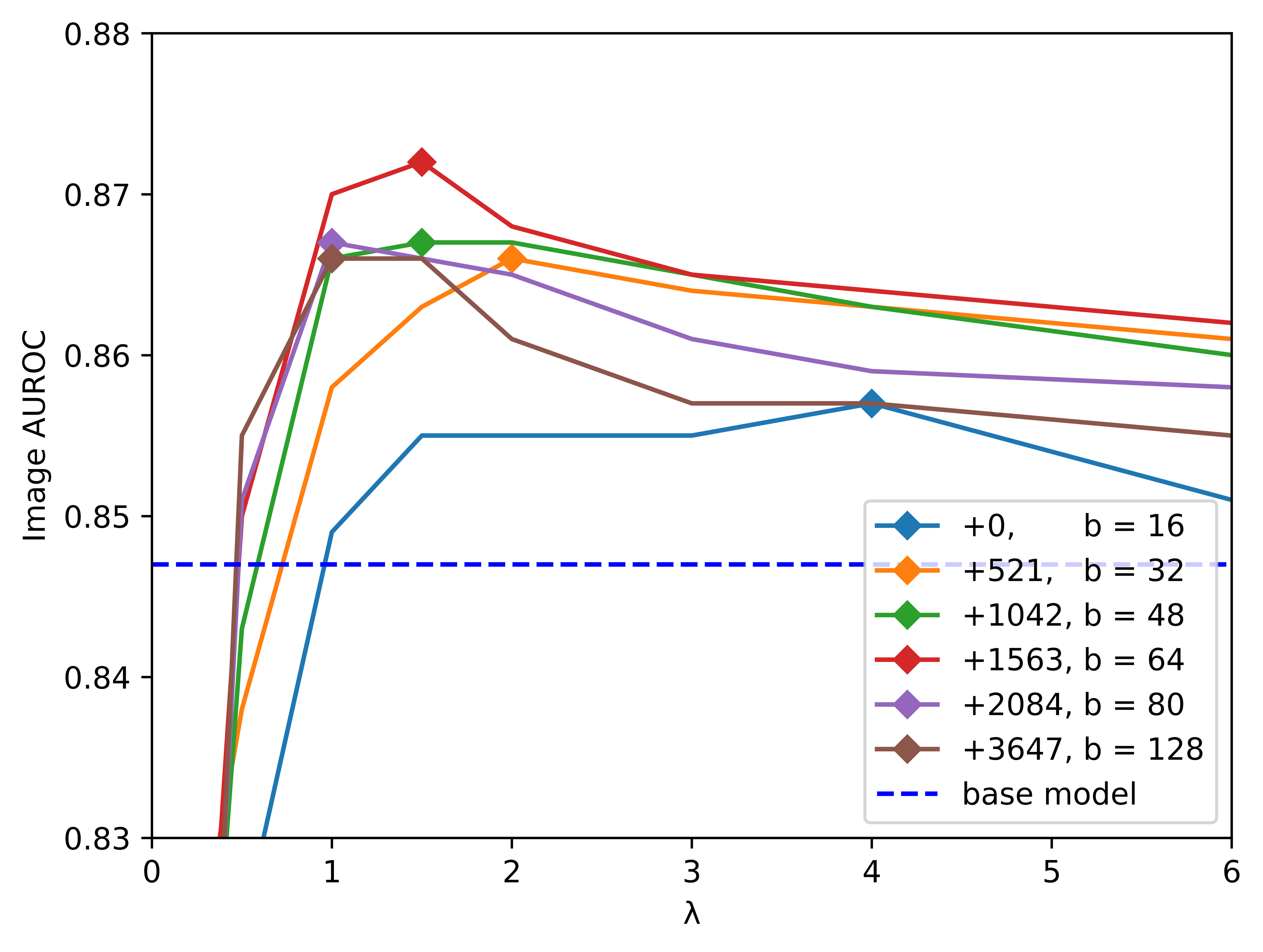}
    AUROC Ablation Results
\end{minipage}
\hfill
% ---------- Right (b): AUPRO ----------
\begin{minipage}{0.48\textwidth}
    \centering
    \includegraphics[width=0.9\linewidth]{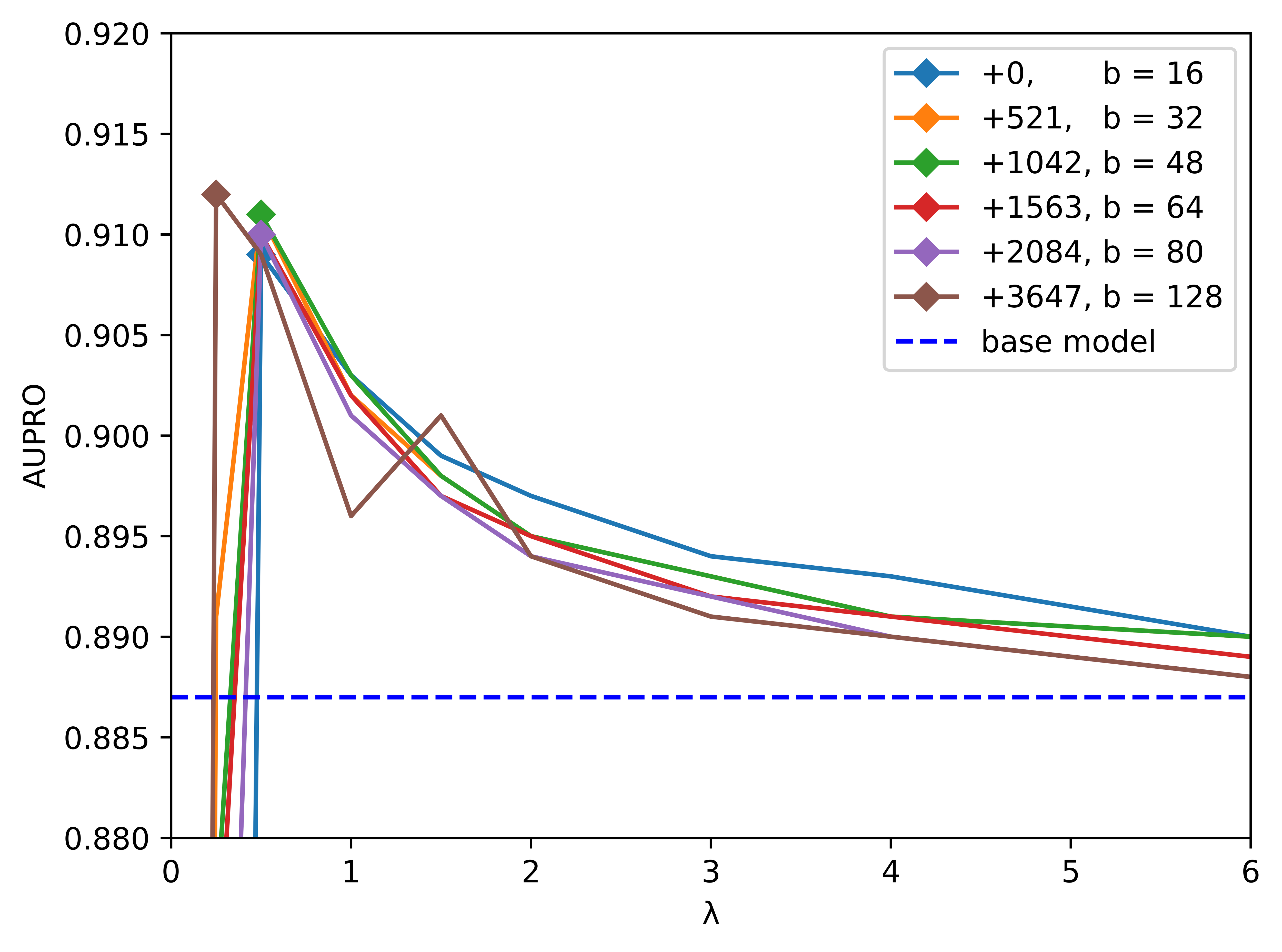}
    AUPRO Ablation Results
\end{minipage}

\vspace{1em}
\caption{Ablation study for the $\lambda$ parameter. '+X' indicates that the training dataset includes the original training data plus $X$ synthetic images. $b$ denotes batch size.}
\label{fig:3way_synth_ablation}
\end{figure*}

In Figure~\ref{fig:3way_synth_ablation}, we ablate the $\lambda$ parameter, which controls the flexibility of the weak teacher 
%(see Section~\ref{sec:method}), 
across varying amounts of synthetic training data. We observe that performance collapses as $\lambda \to 0$ due to lack of regularization, while large $\lambda$ values converge to the base model. Peak performance occurs at intermediate values. Notably, the optimal $\lambda$ decreases as synthetic data increases, suggesting that the synthetic data provides implicit regularization, allowing for greater teacher flexibility without collapse.

%%%%%%%%%%%%%%%%%%%%%%%%%%%%%%%%%%%%%%%%%% FIGURE

%%%%%%%%%%%%%%%%%%%%%%%%%%%%%%%%%%%%%%%%%% FIGURE

%%%%%%%%%%%%%%%%%%%%%%%%%%%%%%%%%%%%%%%%%% FIGURE
\begin{figure*}[t]
    \begin{subfigure}{0.49\textwidth}
        \includegraphics[width=0.9\textwidth]{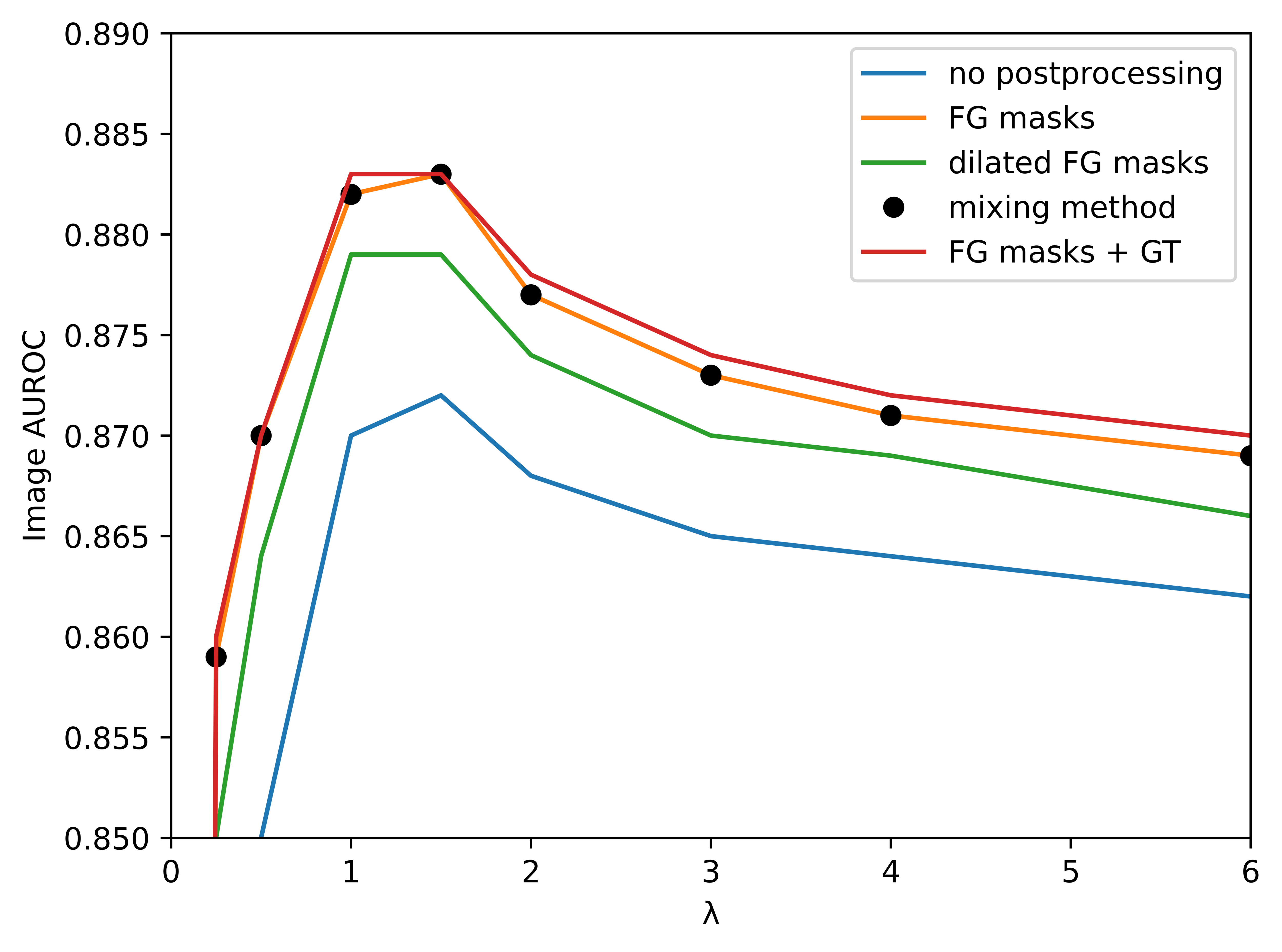}
        \caption{AUROC}
        \label{fig:first}
    \end{subfigure}
    \hfill
    \begin{subfigure}{0.49\textwidth}
        \includegraphics[width=0.9\textwidth]{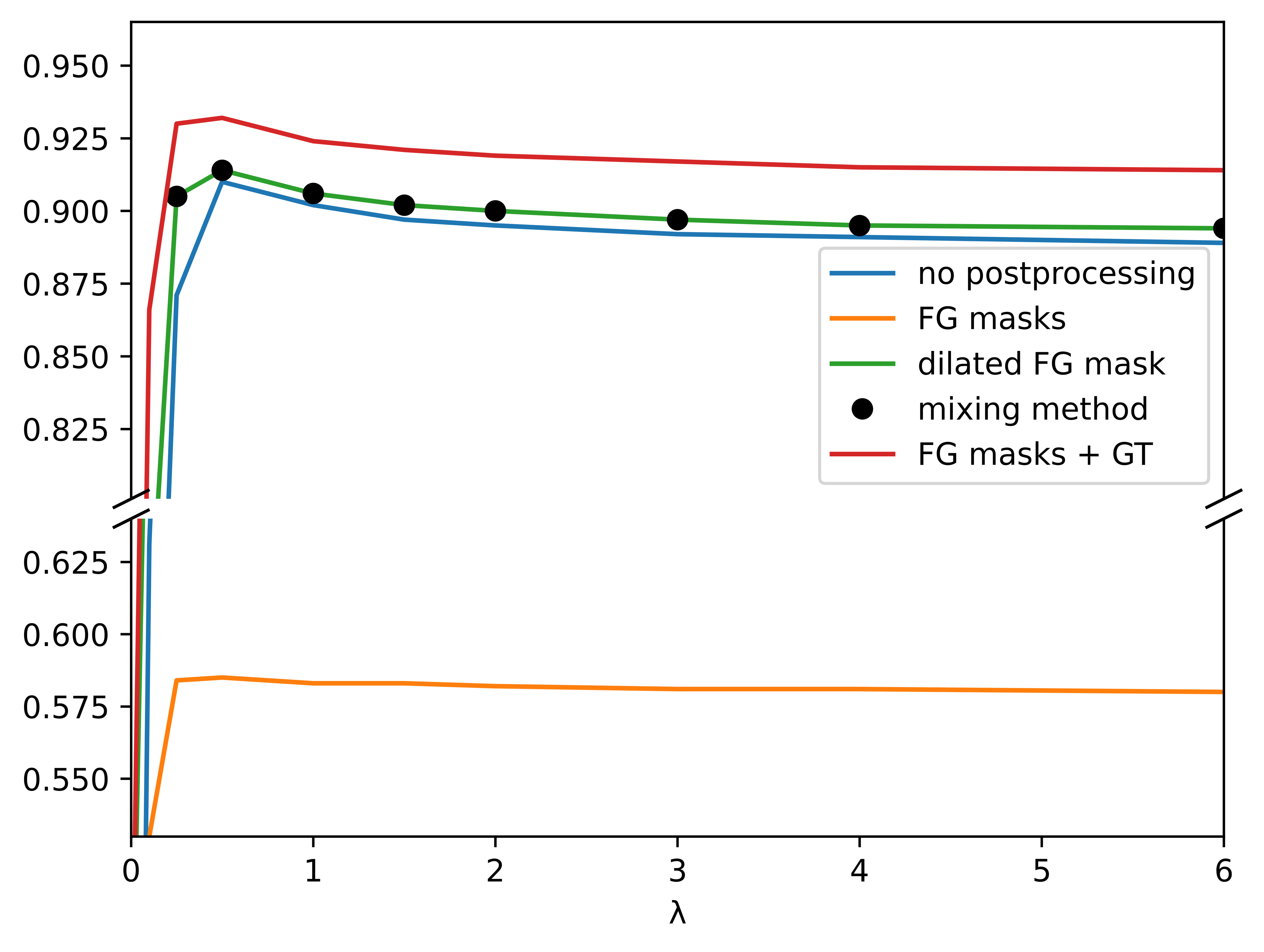}
        \caption{AUPRO}
        \label{fig:second}
    \end{subfigure}
        
\caption{Ablation of the $\lambda$ parameter across multiple scoremap post-processing strategies. 
\emph{FG masks} are the raw Matcher outputs; \emph{dilated FG masks} are FG masks dilated by 40px; 
the \emph{mixing method} fuses FG and dilated FG masks; and \emph{FG masks+GT} take the union of FG masks with ground-truth test masks (upper bound). 
Results are from the final epoch and averaged over five runs.}

\label{fig:3way_synth_ablation_with_postprocessing}
\end{figure*}
%%%%%%%%%%%%%%%%%%%%%%%%%%%%%%%%%%%%%%%%%% FIGURE

%%%%%%%%%%%%%%%%%%%%%%%%%%%%%%%%%%%%%%%%%% FIGURE
\begin{figure}[t]
\centering
\includegraphics[width=\columnwidth]{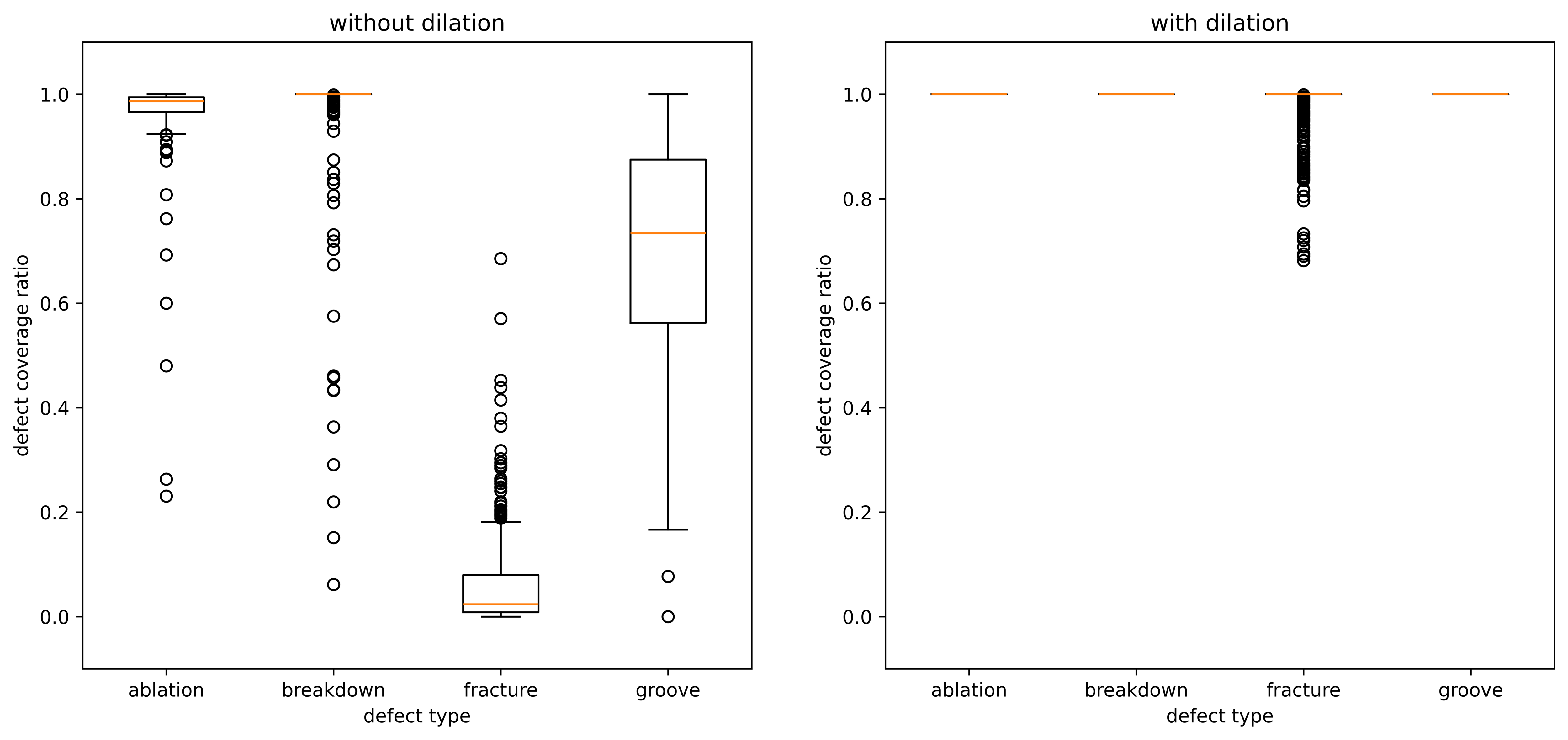}
\caption{Defect coverage ratios of FG masks for the different defect types present in AeBAD-S with and without FG mask dilation.}
\label{fig:defect_ratio_boxplot_AeBAD}
\end{figure}

%%%%%%%%%%%%%%%%%%%%%%%%%%%%%%%%%%%%%%%%%% FIGURE

%%%%%%%%%%%%%%%%%%%%%%%%%%%%%%%%%%%%%%%%%% FIGURE
\begin{figure}[t]
\centering
\includegraphics[width=0.75\columnwidth]{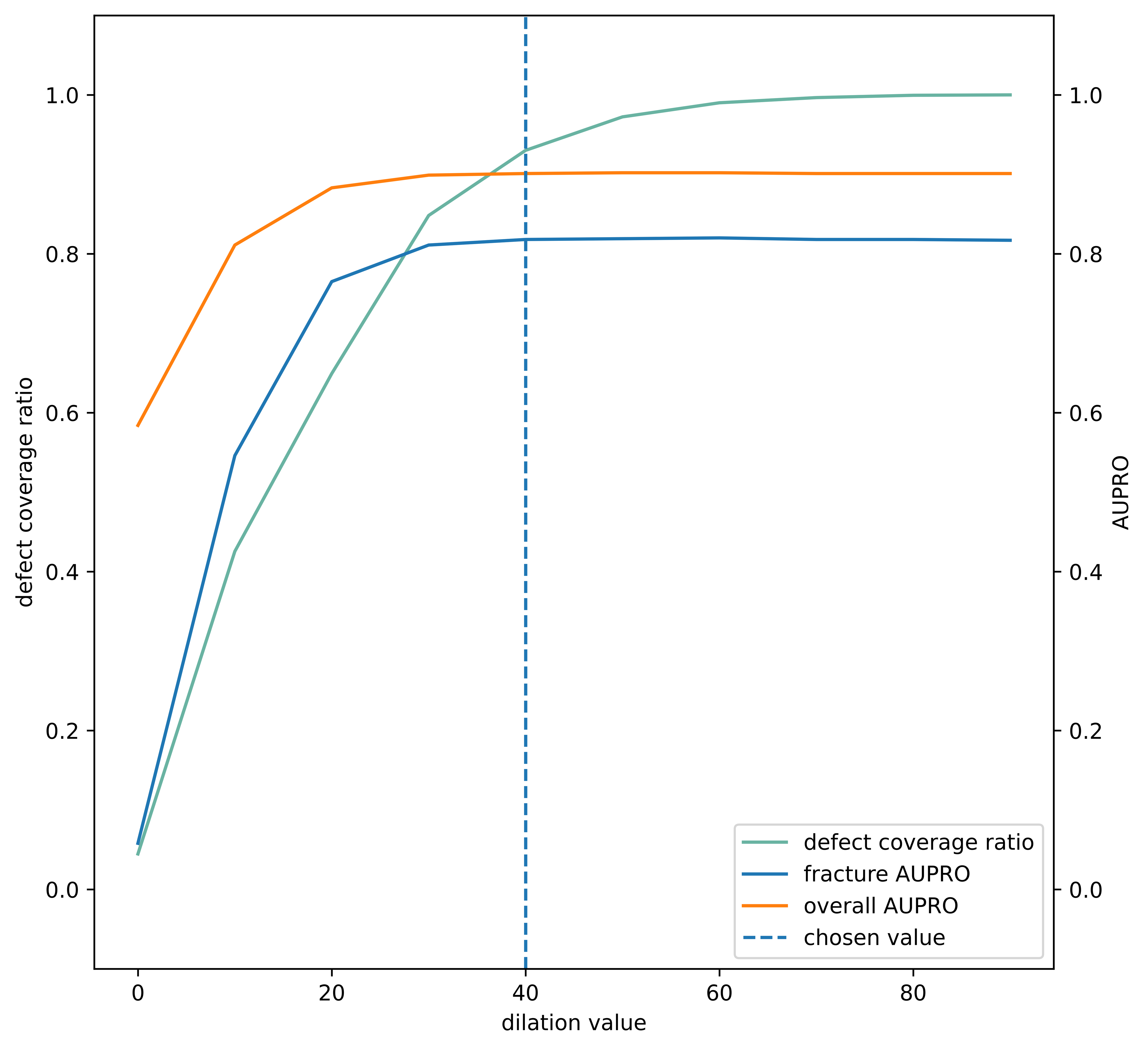}
\caption{Interaction between defect coverage ratio and AUPRO score for defect type 'fracture'}
\label{fig:Defect_ratio_plot_AeBAD}
\end{figure}

\textbf{Background Removal}
Figure~\ref{fig:3way_synth_ablation_with_postprocessing} compares four post-processing variants: 
\emph{FG masks} (raw Matcher output),  
\emph{FG masks+GT} (oracle union with ground truth i.e the \emph{cheated upper bound} which is unavailable at test time),  
\emph{dilated FG masks} (FG masks dilated by 40px), and our \emph{mixing} method that merges raw and dilated masks.  

Raw FG masks already reach the oracle in I-AUROC~(Fig.~\ref{fig:3way_synth_ablation_with_postprocessing}(a)), 
confirming Matcher's accuracy, but their AUPRO collapses because edge and fracture-type defects are missed~(Fig.~\ref{fig:3way_synth_ablation_with_postprocessing}(b)).  
Dilating restores AUPRO at the cost of a slight I-AUROC drop, echoing the defect-coverage curves in 
Figures~\ref{fig:defect_ratio_boxplot_AeBAD}--\ref{fig:Defect_ratio_plot_AeBAD}.  
Our mixing step dilates the FG scoremap and clips it to the peak value of the raw scoremap, retaining sharp discrimination while covering all defects, and thus yields the best overall balance.

\subsection{Ablations on Key Contributions}

To observe how our proposed improvements interact we perform extensive ablation studies on MVTecAD and AeBAD-S. Table~\ref{tab:config_ablation_v2} states the individual and combined effects of our proposed approaches. To assess the quality of the generated FG/BG segmentation masks we compare against a set of manually curated FG/BG masks using Visual Prompting on all images in the test dataset. The results are stated in Table \ref{table:mask_iou_exps}.

\section{Conclusion}
\label{sec:Conclusion}
In this study, we demonstrate that established anomaly detection methods heavily rely on controlled environments, limiting their applicability in real-world scenarios.

%To increase the robustness of methods to variations in object position and background noise we propose a framework which is architecture agnostic and can be applied to most anomaly detection methods. 
To enhance robustness against variations in object positioning and background noise, we propose a framework that is architecture-agnostic and adaptable to most anomaly detection methods.
The framework consists of three independent and modular improvements that are easily applicable:
(1) leveraging a diffusion model to enrich the training dataset with synthetic images, increasing data diversity and enabling the model to capture the data distribution more reliably, thus improving generalization;
(2) unfreezing the teacher encoder in student--teacher reconstruction-based methods, which increases adaptability to the target domain and mitigates pre-training bias, thereby boosting model performance;
(3) employing Visual Prompting to isolate the object of interest by masking the image background, thereby improving robustness to background variation.

Extensive experiments validate the effectiveness of our improvements under challenging conditions. On the AeBAD-S dataset, our framework achieves state-of-the-art performance in both detection and segmentation, improving scores by 3.5\% and 1.1\%, respectively, over the baseline model.

\bibliographystyle{unsrtnat}
\bibliography{aaai2026}

\clearpage

\end{document}